\documentclass[11pt]{article}
\usepackage[utf8]{inputenc}
\usepackage[T1]{fontenc}
\usepackage{graphicx}
\usepackage{longtable}
\usepackage{wrapfig}
\usepackage{rotating}
\usepackage[normalem]{ulem}
\usepackage{amsmath}
\usepackage{amssymb}
\usepackage{capt-of}
\usepackage{hyperref}
\usepackage[final]{latex/acl}
\usepackage{times}
\usepackage{latexsym}
\usepackage[utf8]{inputenc}
\usepackage{microtype}
\usepackage{inconsolata}
\usepackage{enumitem}
\usepackage{multirow}
\date{\today}

\author{
Debarghya Datta \and Soumajit Pramanik \\
Department of Computer Science\\
Indian Institute of Technology, Bhilai\\
\texttt{\{debarghyad,soumajit\}@iitbhilai.ac.in}\\
}

\title{Unsupervised Named Entity Disambiguation for Low Resource Domains}
\begin{document}
\maketitle
\begin{abstract}
In the ever-evolving landscape of natural language processing and
information retrieval, the need for robust and domain-specific entity
linking algorithms has become increasingly apparent. It is crucial in
a considerable number of fields such as humanities, technical writing and
biomedical sciences to enrich texts with semantics and discover more
knowledge. The use of Named Entity Disambiguation (NED) in such domains requires handling noisy
texts, low resource settings and domain-specific KBs.  Existing
approaches are mostly inappropriate for such scenarios, as they either depend on
training data or are not flexible enough to work with domain-specific KBs. Thus in this work, we present an unsupervised approach leveraging
the concept of Group Steiner Trees (GST), which can identify the most relevant candidates for entity disambiguation using the
contextual similarities across candidate entities for all the mentions present in a document. We outperform the state-of-the-art unsupervised methods by more than 40\% (in avg.) in terms of Precision@1 across various domain-specific datasets.

\end{abstract}


\section{Introduction}
Named Entity Disambiguation (NED) is the task of resolving the ambiguity associated
with entity mentions in a document by linking them to the appropriate entries in a
Knowledge Base (KB).
Recently, NED has been applied in various fields, including digital humanities, art, architecture, literature, and biomedical science, for tasks such as searching~\cite{meij2014entity}, question answering
~\cite{yih-etal-2015-semantic} and information
extraction~\cite{nooralahzadeh-ovrelid-2018-sirius}.

The key challenges in such domain specific NED tasks are twofold - 
(a) they provide little or no training data with ground truth annotations and 
(b) the associated knowledge graphs (KG) are typically small and with no or very limited entity descriptions~\cite{shi2023knowledge}. In order to deal with such challenges, in this work we consider the setting where entity disambiguation is needed to be performed with \emph{absolute absence of annotated data}.
In such constrained scenarios, leveraging the state-of-the-art neural entity linkers become infeasible as they are primarily dependent on a 
large corpus of annotated data and long enough entity descriptions from KG~\cite{CadavidSnchez2023EvaluatingEE, arora2021low}.
Similarly this setting also disqualifies unsupervised NED approaches such as ~\cite{pan2015unsupervised} which rely on labeled data to generate candidate entities such as domain-adaptive transformer-based models~\cite{aydin-etal-2022-find}, BLINK~\cite{wu2019zero}, Zeshel~\cite{logeswaran-etal-2019-zero}, and auto-regressive models like GENRE~\cite{decao2021autoregressive}.

In the literature, only a few approaches fit our constrained setting such as graph-based using mention distances~\cite{hoffart2011robust}, PageRank/random walk based~\cite{guo2018robust}, and graph ranking based~\cite{alhelbawy2014graph}. A recent approach by~\cite{arora2021low} also explores singular value decomposition, showing gold entities in a low-rank subspace. However, these methods often struggle in achieving the required efficacy while disambiguating entities.


In this work, we present a novel unsupervised NED approach for domain specific low-resource 
scenarios, which leverages the concept of Group Steiner Trees (GSTs) ~\cite{garg2000polylogarithmic}.
In this approach, we map the candidate entities for each mention in the document, to nodes in the associated knowledge graph, obtain the subgraph connecting these nodes and then extract minimum cost GSTs from this sub-graph. Such GSTs facilitate collective entity disambiguation exploiting the
fact that the entities that are truly mentioned in
a document (the `gold entities') tend to form a
dense subgraph among the set of all candidate entities in the document.


\begin{figure*}
\includegraphics[width=\linewidth]{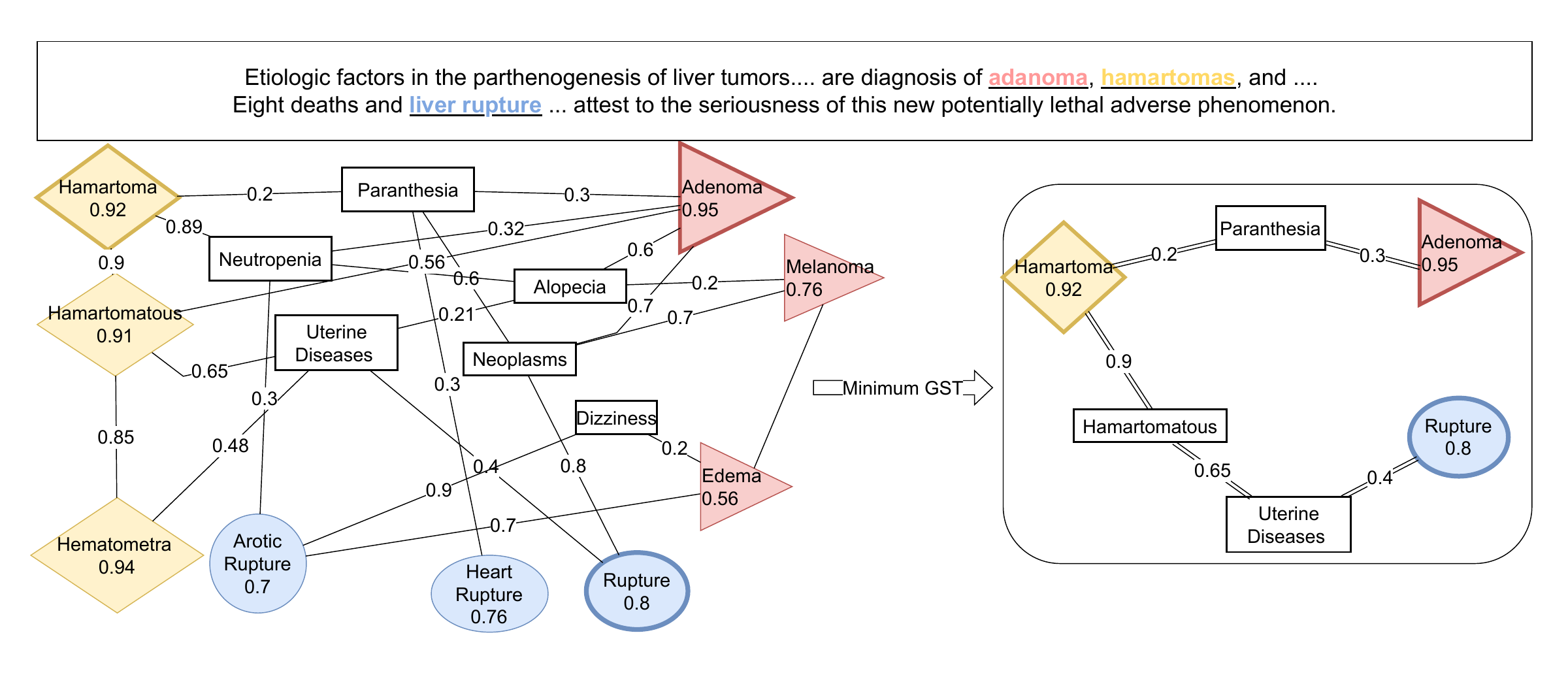}
\caption{Proposed GST-NED approach: The sample document at the top contains three mentions; the subgraph extracted from the KB is shown at the left and the minimum cost GST is shown using the box at the right. In the induced subgraph, the candidates for `adenoma' are marked in red, `hamartomas' are marked in yellow and `liver rupture' are marked in blue.}
\label{fig:el-task}
\end{figure*}

In summary, our main contributions are the following -
(a) We propose an unsupervised \textbf{G}roup \textbf{S}teiner \textbf{T}ree based \textbf{N}amed \textbf{E}ntity \textbf{D}isambiguation (\emph{GST-NED}) method which is capable to perform NED for low resource domains at the absence of any annotated data;
(b) We compare our proposed approach with several state-of-the-art baselines across multiple domain specific datasets and demonstrate its superior performance
with significant improvements in the metrics (more than $40\%$ in avg. in Precision@1 scores)~\footnote{Code is available at \url{https://github.com/deba-iitbh/GST-NED}}. 




\section{Problem Statement}
Similar to most previous works in the NED literature (with a few exceptions \cite{kolitsas2018end, sil2013re}), we assume that document-wise mention spans
(usually obtained by a named entity recognizer) are already
provided.
Let $d$ be a single document from a collection $D$ of documents.
Also, let $M_d = \{m_1, m_2, \ldots , m_M\}$
be the set of $M$ mentions contained in $d$, and let $\mathcal{E}$
be the collection of all the entities contained in the reference domain specific Knowledge Graph $KG$. The task here is to find, for each mention $m_i$ the correct entity $e \in \mathcal{E}$ it refers to.

Typically, given the set of mentions, an NED approach performs the disambiguation in two steps - 
(a) Candidate generation, where candidate entities from the $KG$ are retrieved for each of the mentions, and (b) Candidate Ranking, where the candidate entities are ranked based on their propensities to be mapped with the corresponding mentions. 
Our primary focus in this study is the candidate ranking/disambiguation step. 
In the following, we describe our proposed candidate ranking method and mention the approaches adhered for the other step.

\section{Methodology}
\subsection{Candidate Generation}
We index the domain specific $KG$ and use fuzzy text search \cite{max_bachmann_2021_5584996} to retrieve candidates based on the surface form of the annotated mention. This is found to be the standard practice in most of the recent unsupervised NED approaches~\cite{yang2023b, simos2022computationally}
Fuzzy text search returns a confidence value with each potential match; we keep only the candidates which are returned with more than $0.75$ confidence value (chosen empirically)~\footnote{In case of exact match with a KG node, we consider it to be the correct match for the mention and skip the candidate ranking step.}.

\subsection{Candidate Ranking}

We use the knowledge graph ($KG$) to create a subgraph connecting all pairs of candidate entities obtained from the candidate generation step for a particular document $d$. To keep the graph size manageable, we limit path lengths to be a maximum of three hops between entity candidates. We further enhance the graph by adding node weights based on the Jaro-Winkler distance~\cite{wang2017efficient} (reflecting similarities of candidates with mentions), and edge weights based on cosine similarities of Node2Vec \cite{grover2016node2vec} structural embeddings of the endpoints. 
In Figure~\ref{fig:el-task}, we depict a document with three mentions and the corresponding induced subgraph of candidate entities (left side).

\noindent
\textbf{Finding GST:} Our approach to identify the correct candidates
relies on the intuition that a gold entity candidate from a document $d$ should be more tightly connected with other gold candidates in the induced subgraph   
compared to other non-gold candidates.
In other words, we expect the gold entities within the induced subgraph to form cohesive and closely linked subgraphs due to their contextual proximities (as they are used in the same document).
In order to exploit this intuition, we first define the notion of terminals - for every mention $m_i$, we denote the corresponding candidate entity nodes as the terminal nodes for that mention and group them together as $T_i$. 
Further the task remains is to select the correct candidate node from each terminal group for which we leverage the concept of Group Steiner Trees (GST)~\cite{ding2006finding,pramanik2024uniqorn} as defined below,
\begin{itemize}[topsep=2pt,itemsep=0pt,partopsep=0pt, parsep=0pt]
\item Given an undirected and weighted graph \((V, E)\) and given groups of terminal nodes \(\{T_1, . . . ,T_l\}\) with each \(T_\nu \subseteq V\), compute the minimum-cost tree \((V^*, E^*)\) that connects at least one node from each of \(\{T_1, . . . ,T_l\}\): \(\min \sum_{ij \in E^*} c_{ij}\) s.t. \(T_\nu \cap V^* \ne \emptyset, \ \forall T_\nu\).
\end{itemize}
In our case, we consider $c_{ij}=(1 - w_{ij})$ where $w_{ij}$ represents the edge weight between nodes $i$ and $j$. 
As per definition, each GST would have to necessarily choose at least one candidate entity from each of the terminal groups. Hence, each detected GST would provide at least one potential solution to the entity disambiguation problem. 
As we further posit that the gold candidate entities are more tightly connected compared to non-gold candidates, the probability of the gold candidates to be chosen in the minimum cost GST increases (as the minimum cost GST ensures shorter distances between the chosen candidates and higher weighted edges i.e. lower edge-costs). For instance, in the right side of the Fig.~\ref{fig:el-task}, we depict that the minimum cost GST extracted from the induced subgraph contains all the gold candidate entities corresponding to the mentions in the document.

\noindent
\textbf{Relaxation to GST-k and Ranking Criteria}: 
In our setting, we actually look for the entity candidates extracted from k least cost GSTs (used k=10 for our work empirically) rather than relying upon only the minimum cost GST. This is for enhancing the robustness of the approach as it allows us to rank the different candidate entities efficiently.
We utilize the following three intutive ranking schemes to rank the candidate entities for each mention and choose the higher ranked one -  \textbf{(a) GST count}: Number of GSTs where the candidate is present; the higher the better, \textbf{(b) GST Cost}: Total cost of the GSTs where the candidate is present; the lower the better, and
\textbf{(c) Node Weight:} The sum of node weights in the GSTs where the candidate is present; the higher the better. Subsequently, we compare the performance of all three schemes to choose the best one.

\noindent
\textbf{Complexity:} Steiner trees are among the classical NP-complete problems~\cite{ding2006finding}, and this holds for the GST problem too. However, the problem has tractable fixed-parameter complexity when the number of terminals is treated as a constant \cite{downey2013fundamentals}, and there are also good polynomial-time approximation algorithms extensively applied in the area of keyword search over databases \cite{ding2006finding,kacholia2005bidirectional,li2016efficient}. In
\emph{GST-NED}, we build on the exact solution method by \cite{ding2006finding}, which uses a dynamic programming approach and has exponential runtime in the number of mentions (which is typically limited) but has $O(n\log n)$ complexity in the graph size.

\section{Experimental Setup}

\subsection{Datasets}
In order to show the efficacy of our model, we choose the following four datasets from diverse domains of literature, law, museum artifacts and chemicals (see Table.~\ref{tab:data-stat} for more details). 

\noindent
\textbf{WWO}\footnote{\url{https://www.wwp.northeastern.edu/wwo}} is a collection of textual documents (poems, plays and novels) by pre-Victorian women writers, partially
annotated~\cite{flanders2010encoding} with person, works and places entities. 

\begin{table}
\resizebox{\columnwidth}{!}{
\begin{tabular}{lrrrrrr}
Dataset & \#D & \#M & \#N & \#E & \#C & \#R\\[0pt]
\hline
WWO & 76 & 14651 & 9065 & 4936 & 10 & 0.83\\[0pt]
1641 & 16 & 480 & 3503 & 338 & 10 & 0.26\\[0pt]
Artifact & 168 & 6311 & 41180 & 42634 & 11 & 0.66\\[0pt]
Chemical & 135 & 15769 & 176415 & 249275 & 10 & 0.73\\[0pt]
\end{tabular}
}
\caption{\label{tab:data-stat}Data statistics of the four used datasets: Total number of Documents ($\#D$), Total number of mentions ($\#M$), Number of Nodes ($\#N$) and Edges ($\#E$) in KG, Average number of candidates per mention ($\#C$) and Recall of the candidate entities i.e fraction of mentions with gold entities present among the candidates ($\#R$)}.
\end{table}

\noindent
\textbf{1641}\footnote{\url{http://1641.tcd.ie/}} consists of legal texts in the form of court witness statements recorded after the Irish Rebellion of 1641, partially
annotated with person names against a subset of DBpedia KB~\cite{klie2020zero}. 

\noindent
\textbf{Chemical} dataset 
is sourced from the BC5CDR corpus ~\cite{li2016biocreative}. It features a comprehensive human annotations of chemicals, each tagged with unique MeSH identifiers. 
For the categorization of chemicals, the Chemicals vocabulary is sourced from the Comparative Toxicogenomics Database (CTD)~\footnote{\url{https://www.ctdbase.org/}}. 

\noindent
\textbf{Artifact}~\cite{CadavidSnchez2023EvaluatingEE} is a collection of digital descriptions of Museum objects 
annotated with four different text fields:
title, detailed description, free-form metadata against the Getty
Arts, and Architecture Thesaurus~\footnote{\url{https://www.getty.edu/research/tools/vocabularies/aat/about.html}}(AAT). 




\subsection{Baselines}
To compare the performance of our proposed approach, we leverage the following baselines~\footnote{All the Datasets and Baseline codes are available under MIT \& Apache License.}.

\noindent
\textbf{NameMatch}\cite{klie2020zero}. We employ a string-matching approach to select candidates that exactly match the surface form of the mention.

\noindent
\textbf{BLINK*}~\cite{wu2019zero}. We adapt a fine-tuned BLINK model in our domain specific setup for predicting named entities for each mention. As it matches entities to Wikipedia\footnote{\url{https://www.wikipedia.org/}} by default, we subsequently perform a fuzzy matching process to align the predicted entities with our domain specific knowledge base.


\noindent
\textbf{WalkingNED}~\cite{guo2018robust} is a graph-based approach to disambiguate the mention candidates, based on local similarity (surface form similarity) and global similarity (similarity between the semantic signatures of the candidate and the document computed using PageRank). 


\noindent
\textbf{Eigenthemes}~\cite{arora2021low} 
is an approach which
leverages the inherent property of `gold entities' to cluster together within the embedding space by representing entities as vectors and utilizing Singular Value Decomposition (SVD).

\begin{table}[htbp]
\centering
\begin{tabular}{p{2cm}p{2cm}p{1cm}p{1cm}}
Dataset & Model & P@1 & HIT@5\\[0pt]
\hline
 WWO
 & NameMatch	& 0.35 & 0.35 \\
 & BLINK* & 0.07 & 0.09 \\
 & WalkingNED & 0.18 & 0.49\\
 & EigenThemes & 0.14 & 0.45\\
 & GST-NED & \textbf{0.57} & \textbf{0.72}\\
\hline
 1641
 & NameMatch & 0.06 & 0.06 \\
 & BLINK* & 0.05	& 0.11 \\
 & WalkingNED & 0.11 & 0.17\\
 & EigenThemes & 0.17 & \textbf{0.25}\\
 & GST-NED & \textbf{0.20} & 0.22 \\
\hline
Artifact
 & NameMatch & 0.23 & 0.23 \\
 & BLINK* & 0.02	& 0.03 \\
 & WalkingNED & 0.26 & 0.56\\
 & EigenThemes & 0.15 & 0.44\\
 & GST-NED & \textbf{0.54} & \textbf{0.61} \\
\hline
 Chemical  
 & NameMatch	& 0.08 & 0.08 \\
 & BLINK* & 0.13 & 0.22 \\
 & WalkingNED & 0.50 & \textbf{0.66}\\
 & EigenThemes & 0.36 & 0.59\\
 & GST-NED & \textbf{0.52} & \textbf{0.66}\\
 \hline
\end{tabular}
\caption{\label{tab:org1c6e60d} NED performance comparison for WWO, 1641, Artifact and Chemical datasets.}
\end{table}


\subsection{Metrics}
Similar to the state-of-the-art literature in NED, we use Precision@1 (correctness of top ranked candidate) and Hit@5 (presence of gold entity in top five ranked candidate) as our evaluation metrics.

\section{Results and Discussion}
We compared our proposed \emph{GST-NED} approach with other baselines algorithms and the corresponding results are depicted in Table.~\ref{tab:org1c6e60d}. We can observe that our method outperforms the state-of-the-art in all the datasets (especially in terms of $P@1$). In 1641, the relatively poor performance of all the algorithms stems from the poor recall of the candidate entities (see Table.~\ref{tab:data-stat}). BLINK* in general works poorly as it struggles to find a suitable match in the domain specific knowledge bases.

\begin{table}[htbp]
    \centering
    \begin{tabular}{c|c|c}
     & WWO	& Artifact \\
    \hline
    GST count	& \textbf{0.57} & \textbf{0.54} \\
    GST cost	& 0.55 & 0.51 \\
    Node weight	& 0.54 & 0.53 \\
    \end{tabular}
    \caption{Comparison over ranking schemes (P@1) on two datasets}
    \label{tab:abl_tab}
\end{table}

\noindent
\textbf{Analysing Ranking Schemes}
 In Table.~\ref{tab:abl_tab}, we analyse the impact of choosing different ranking schemes for candidate ranking in $\emph{GST-NED}$. It is observed that the GST-count scheme performs the best in our scenario.

\noindent
\textbf{Parameter Fine-tuning}
In order to optimize the metric values, we conduct extensive empirical experiments with varying fuzzy threshold values for candidate generation and different numbers of top-ranked GSTs (k) for candidate ranking. These experiments are performed on a small held-out subset ($~10\%$) of the 'WWO' and 'Artifact' datasets, with results presented in Table.~\ref{tbl:hyper_threshold}, \ref{tbl:hyper_k}.
Based on our analysis, considering the fuzzy threshold value of $0.75$ and top-10 GSTs yield the highest Precision@1 score for our setup. Consequently, these parameters are used for all the experiments reported in this work.
 
\begin{table}[htbp]
\centering
\begin{tabular}{c|cc}
Threshold & WWO & Artifact \\ \hline
0.70 & 0.632 & 0.574 \\
0.75 & 0.634 & 0.580 \\
0.80 & 0.632 & 0.562 \\
0.85 & 0.631 & 0.554 \\
0.90 & 0.633 & 0.554 \\
\end{tabular}
\caption{\label{tbl:hyper_threshold}Precision@1 for held-out WWO and Artifact datasets with various Fuzzy Matching thresholds}
\end{table}

\begin{table}[htbp]
\centering
\begin{tabular}{c|cc}
k & WWO & Artifact \\ \hline
1  & 0.63 & 0.55 \\
5  & 0.63 & 0.57 \\
10 & 0.64 & 0.58 \\
20 & 0.62 & 0.56 \\
50 & 0.63 & 0.56 \\
\end{tabular}
\caption{\label{tbl:hyper_k}Precision@1 for held-out WWO and Artifact datasets at various k (number of top ranked GST) values}
\end{table}
\noindent
\textbf{Error Analysis}
We conduct a detailed error analysis to identify the distribution of errors in our proposed pipeline. Specifically, we compute the proportion of instances where error occurs due to: (a) the gold (correct) entity not being present in the candidate list, (b) the gold entity being present in the candidate list but not in the top-k GSTs, and (c) the gold entity being included in the top-k GSTs but does not rank in the top-1 position. On the ‘WWO’ dataset, 14\% of errors corresponded to (a), 11\% to (b), and 18\% to (c), while the remaining 57\% of cases were correctly resolved, resulting in a precision@1 score of 0.57. These findings suggest that enhancing both the ranking mechanism and candidate generation process are critical for achieving improved performance.

\section{Conclusion}
In this paper, we have addressed the problem of NED of domain-specific corpora in the absence of annotated data. It works based on the intuition that a gold
entity candidate from a document should be more cohesively connected with other gold candidates in the
knowledge graph compared to other non-gold candidates. We have leveraged the concept of Group Steiner Trees (GSTs), that relies solely on the availability of candidate entity names and a domain specific knowledge graph. Extraction of minimum cost GSTs in our proposed approach \emph{GST-NED}, ensures that the chosen entities are closely connected in the domain specific knowledge graphs.
Experiments on benchmark datasets from varied domains have portrayed the effectiveness of our proposed approach against the state-of-the art unsupervised and zero-shot approaches.

\section*{Limitations}
Our entity disambiguation method, \emph{GST-NED}, depends on the presence of sufficient number of entities per document to function accurately as we rely upon joint disambiguation of entities. As a result, when the entity count is very low, it fails to provide the correct response.
On the other hand, considering relatively longer document chunks with too many entities increases the graph size, affecting our computational efficacy. Hence, it is essential to analyze this trade-of with a detailed and thorough study. Interestingly, considering longer documents also enhances the possibility of same mention being used multiple times with different meanings which is beyond the capability of our model for the time being. 
Additionally, further works need to be done to improve the scalability of the Steiner tree algorithm we use to compute the optimal trees. Presently it takes around 2 seconds per document for small KGs like WWO, 1641 or Artifact and around 40 seconds per document on the relatively larger KG of Chemical dataset (on a system with $3.9$GHz CPU with $16$ GB RAM).


\noindent

\section*{Ethics}
The data and models in this work are publicly
available. They could contain bias, and should be
used with discretion.

\bibliography{acl_latex}


\end{document}